\documentclass{Interspeech}

\usepackage{comment}

\interspeechcameraready

\title{Meta-Learning Approaches for Speaker-Dependent Voice Fatigue Models}

\author[affiliation={1}]{Roseline}{Polle}
\author[affiliation={1}]{Agnes}{Norbury}
\author[affiliation={1,2}]{Alexandra Livia}{Georgescu}
\author[affiliation={1,2}]{Nicholas}{Cummins}
\author[affiliation={1}]{Stefano}{Goria}

\affiliation{thymia}{London}{UK}
\affiliation{Institute of Psychiatry, Psychology \& Neuroscience}{King's College London}{UK}
\email{roseline@thymia.ai,stefano@thymia.ai}

\begin{document}

\maketitle

\begin{abstract}
Speaker-dependent modelling can substantially improve performance in speech-based health monitoring applications. While mixed-effect models are commonly used for such speaker adaptation, they require computationally expensive retraining for each new observation, making them impractical in a production environment. We reformulate this task as a meta-learning problem and explore three approaches of increasing complexity: ensemble-based distance models, prototypical networks, and transformer-based sequence models. Using pre-trained speech embeddings, we evaluate these methods on a large longitudinal dataset of shift workers (N=1,185, 10,286 recordings), predicting time since sleep from speech as a function of fatigue, a symptom commonly associated with ill-health. Our results demonstrate that all meta-learning approaches tested outperformed both cross-sectional and conventional mixed-effects models, with a transformer-based method achieving the strongest performance. 
\end{abstract}

\section{Introduction}
Fatigue is a common symptom in many physical and mental health conditions and is especially widespread in occupational settings known for high rates of health problems, such as shift work \cite{menting2018fatigue, brown2020mental}. Monitoring fatigue through readily available devices such as smartphones could enable earlier intervention and better outcomes through more targeted treatment adjustments and timely rest provision in safety-critical contexts \cite{adao2021fatigue,karvekar2021smartphone}.

Recent work has shown that fatigue manifests detectably in speech, offering a scalable, low-burden monitoring approach that can be deployed in various real-world settings \cite{huckvale2020prediction,vekkot2024continuous}. Studies have demonstrated that speaker-specific modelling substantially improves detection performance \cite{norbury2024predicting}, likely by better separating speaker characteristics (e.g., age, sex) from fatigue-related changes. However, current speaker adaptation methods such as mixed-effects modelling are impractical for production environments, as they require computationally expensive retraining for each new observation \cite{laird1982random}.  While solutions such as federated learning could distribute this computational load, it would still require complex server-client coordination and would face performance constraints on edge devices \cite{lim2020federated}. An ideal monitoring solution would require users to report their fatigue levels during an initial calibration period, after which the system could maintain accurate predictions from speech alone without retraining.

Meta-learning offers a promising framework to achieve this goal while remaining tractable for real-world deployment \cite{hospedales2021meta}. Under this `learning to learn' paradigm, models learn to leverage a few examples, known as a support set, from initial user interactions to make predictions about new speech samples, without requiring further user input or model retraining \cite{li2018learning}. Meta-learning has been demonstrated for healthcare applications such as rare disease diagnosis \cite{rafiei2024meta, singh2023meta}, and for speech processing tasks including emotion classification \cite{chopra2021meta}.

In this work, we reformulate fatigue monitoring as a meta-learning problem and explore three approaches of increasing complexity: (i) a simple ensemble-based distance model that learns changes between speech samples, (ii) a prototypical network that learns a metric space for comparing speech samples \cite{snell2017prototypical}, and (iii) A transformer-based sequence model that learns to predict fatigue from sequences of previous observations \cite{yadlowsky2023pretrainingdatamixturesenable}. 

Using a large longitudinal dataset of shift workers \cite{norbury2024predicting}, we compare our meta-learning approaches against two baselines: (i) a cross-sectional model that ignores speaker identity, and (ii) a linear mixed-effects model \cite{oberg2007linear}. Results demonstrate that our meta-learning methods outperform both baselines after collecting just a few user-reported fatigue levels. The transformer-based approach performs strongest while offering practical advantages for deployment; it requires no speaker-specific retraining, uses a single model artefact, and handles both classification and regression naturally.

\section{Data and Preprocessing}
\label{section:data}
\subsection{Dataset}
Our analysis uses data from a large-scale longitudinal study of shift workers \cite{norbury2024predicting}, comprising 1,185 participants monitored over a two-week period. Study procedures were reviewed by an independent research ethics expert (Dr. David Carpenter) working under the auspices of the Association of Research Managers and Administrators (ARMA, https://arma.ac.uk/, involved in formulating national standards for research ethics support and review) and received a favourable ethics opinion on 22nd November 2023. All participants gave written informed consent. Access to the dataset can be obtained through reasonable requests to the senior author.

During the study, participants completed speech recordings and self-report assessments twice daily on their personal devices, coinciding with the start and end of their work shifts. From the collected metrics, we focus on time since sleep ($S$) as our target variable, representing a key component of fatigue \cite{caldwell2019fatigue}. Among the various speech tasks completed by participants, we analysed the reading of fourteen different texts from the speech literature, including the Aesop fables, The North Wind and the Sun \cite{international1999handbook}, The Boy who Cried Wolf \cite{deterding_north_2006}, and the Rainbow Passage \cite{fairbanks_voice_1960}. On each day, different texts (but the same for all participants, although translated into the relevant language, i.e. English or Spanish), were assigned before and after each work shift. These standardized reading tasks provide a controlled context where variations in speech primarily reflect paralinguistic features rather than content or speaking style \cite{adam_2020_phonetic}.

\subsection{Data Cleaning and Processing}
\begin{figure}[t]
  \centering
  \includegraphics[width=\linewidth]{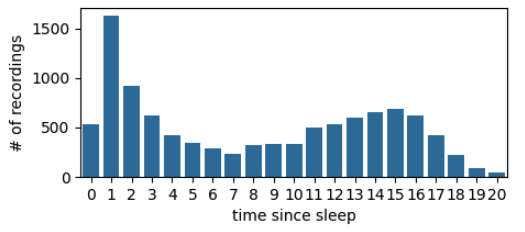}
  \caption{Time since sleep samples distribution}
  \label{fig:tss_distrib}
\end{figure}

The dataset underwent several preprocessing steps to ensure quality and consistency. For the target variable, time since sleep values were constrained to the range [0,24] hours, resulting in 10,286 valid observations across 1,185 speakers. The distribution of these values shows expected bimodality with peaks at 1 and 15 hours (Figure \ref{fig:tss_distrib}), reflecting the pre/post-shift recording schedule. To support regression and classification analysis, we additionally defined a binary 'fatigued' state where participants with time since sleep greater than or equal to 10 hours were labelled as fatigued. This threshold was selected based on prior research in shift work environments \cite{norbury2024predicting} and results in a reasonably balanced dataset with 45.4\% of the samples labelled as fatigued. The regression task thus aims to predict the continuous time since sleep value, while the classification task predicts this binary fatigue state.

For the audio data, all recordings were standardized to WAV format with 16kHz sampling rate, 32-bit resolution, and mono-channel audio. From each recording, leading and trailing silence were removed, and then, to maintain a standardized approach, the first 20 seconds were selected for feature analysis. Speech embeddings were then generated using Trillsson-5 \cite{shor2022trillsson}, a model specifically trained to capture paralinguistic information. This approach produces fixed-dimension feature vectors that encapsulate speaker state, while being robust to variations in recording conditions \cite{shor2022conformers}.

The final processed dataset characteristics are detailed in Table \ref{tab:data}. The dataset's size and diversity, combined with its longitudinal nature, make it particularly suitable for evaluating speaker-dependent modelling approaches.

\begin{table}[t]
  \caption{Dataset Characteristics}
  \label{tab:data}
  \vspace{-3mm}
  \footnotesize 
  \setlength{\tabcolsep}{3pt}  % reduce column separation
  \footnotesize  % reduce font size
  \centering
  \begin{subtable}{\columnwidth}
    \centering
    \caption{Demographics}
    \vspace{-2mm}
    \begin{tabular}{@{}lrrr@{}}
      \toprule
      \textbf{Group} & \textbf{Participants} & \textbf{\%} & \textbf{avg samples} \\
      \midrule
      \textbf{Total} & 1,185 & 100.0 & 8.7 \\
      \midrule
      \textbf{Gender} & & & \\
      Female/Male/Other & 553/528/4 & 55.1/44.6/0.3 & 8.2/9.3/4.2 \\
      % Female & 553 & 55.1 & 8.2 \\
      % Male & 528 & 44.6 & 9.3 \\
      % Prefer not to say & 4 & 0.3 & 4.2 \\
      \midrule
      \textbf{Age} & & & \\
      Under 40/ Over 40 & 936/250 & 78.9/21.1 & 8.4/9.6 \\
      % Under 40 & 936 & 78.9 & 8.4 \\
      % Over 40 & 250 & 21.1 & 9.6 \\
      \midrule
      \textbf{Language} & & & \\
      GB/US/es/eng-other & 701/268/170/46 & 59.2/22.6/14.3/3.9 & 8.6/7.1/12.1/6.0 \\
      % GB & 701 & 59.2 & 8.6 \\
      % en-US & 268 & 22.6 & 7.1 \\
      % es & 170 & 14.3 & 12.2 \\
      % en-OTHER & 46 & 3.9 & 6.0 \\
      \bottomrule
    \end{tabular}
  \end{subtable}
  
  \vspace{1mm}
  
\begin{subtable}{\columnwidth}
  \centering
  \caption{Target Variable Characteristics}
  \vspace{-2mm}
  \begin{tabular}{l r l}
    \toprule
    \textbf{Characteristic} & \textbf{Value} & \textbf{Distribution} \\
    \midrule
    Total Samples & 10,286 & (100\%) \\
    Time Since Sleep ($S$ - hours) & 8[6] & avg[std] \\
    Fatigued ($S \geq 10$) & 4,671 & (45.4\%) \\
    Not fatigued ($S < 10$) & 5,615 & (54.6\%) \\
    \bottomrule
  \end{tabular}
\end{subtable}
\vspace{-4mm}
\end{table}

\section{Methods}
\label{section:methods}
Given a speech embedding $x^i_j \in \mathbb{R}^d$ from speaker $i$ at time $j$, our goal is to predict the associated fatigue level $y^i_j$. The target $y^i_j$ can represent either time since sleep (regression) or a binary fatigued/non-fatigued state (classification). Each speaker $i$ provides a sequence of ordered observations ${(x^i_j, y^i_j)}$ with $j=0,...,T$. We aim to learn a model $f$ that can adapt to speaker-specific patterns. At time $t$, the model makes predictions using both the current speech sample $x^i_t$ and a support set  $S^i_t = {(x^i_j, y^i_j)}$ with ${j<t}$ containing all previous observation pairs from that speaker. The support set grows as more observations become available, allowing the model to better adapt to speaker-specific patterns.

We explore three meta-learning approaches of increasing complexity, comparing them against simpler benchmarks (mixed-effects and cross-sectional models that ignore speaker identity) on both regression and classification tasks. 

\subsection{Training and Evaluation}
All models were evaluated using a 5-fold cross-validation scheme with speaker stratified 70\%/10\%/20\% train/validation/test splits; except for the mixed-effects baseline in which we replicated the evaluation procedure in \cite{norbury2024predicting}. To account for potential sequence ordering effects, we created five random sequence orderings (as per section \ref{subsection:sensitivity_analysis}) for each speaker within each cross-validation split, resulting in 25 total evaluation iterations. We report AUC, balanced accuracy, precision, and recall for classification tasks. We compute Pearson correlation coefficients and root mean squared error for regression tasks. All reported metrics are averaged across the 25 iterations, with standard deviations indicating performance stability across different sequence orderings and data splits. For classification models, decision thresholds were optimised on the validation set to equalise precision and recall, ensuring balanced performance across classes without test set leakage. 

\subsection{Benchmark Models}

\noindent
\textbf{Cross-sectional Model:} Our simplest baseline ignores speaker identity and treats each observation independently, learning a mapping $f_{cs}: x \rightarrow y$ directly from speech embeddings to fatigue levels. For regression, we use ridge regression \cite{hoerl1970ridge}, with a parameter search for the regularisation parameter $\alpha$ in $(0.1, 1, 10, 100,1000)$ . For classification, we use logistic regression, with parameter search for the inverse regularisation strength $C$ in $(0.001,0.01,0.1,1,10)$. Parameters are tuned on the validation set.

\noindent
\textbf{Mixed-effects Model:} We implement a mixed-effects approach \cite{oberg2007linear}, following \cite{norbury2024predicting} to incorporate speaker-specific patterns (with fixed effects for age, sex, and language, and random effects for participants). For each speaker $i$ and time point $t$, we train a separate ridge regression model using all previous observations ${(x^i_j, y^i_j)}_{j<t}$ from that speaker. Unlike our meta-learning models, which can be trained once and then adapted to new speakers, this approach requires full model retraining with each new observation, making it considerably less practical in deployment. The evaluation methodology for this baseline differs from our other models: following \cite{norbury2024predicting}'s approach on the same dataset, we use speakers' past observations as training data and their current observations as test data. As this model serves as a benchmark for comparison with our proposed methods, we deem this methodological difference acceptable.

\subsection{Meta-learning Approaches}

\noindent
\textbf{Distance-based Model:} Our first meta-learning approach focuses on learning changes in fatigue levels rather than absolute values. By analyzing fatigue-related variations in speech within each individual, this framework inherently accommodates speaker-specific differences. Instead of predicting fatigue directly, we train a ridge regression model ($\alpha=1000$) to predict differences:
\begin{equation}
\Delta y = f_d(x_j - x_t)
\end{equation}
For a new observation $x^i_t$, we predict:
\begin{equation}
y^i_t = \begin{cases}
f_{cs}(x^i_t) & \text{if } t = 0 \\
\frac{1}{|S^i_t|}\sum_{j \in S^i_t}(y^i_j + f_d(x^i_t - x^i_j)) & \text{otherwise}
\end{cases}
\end{equation}
where $S^i_t$ is the set of indices of previous observations for speaker $i$. Note that this approach requires two models: the cross-sectional model $f_{cs}$ for the initial prediction and the distance model $f_d$ for subsequent predictions based on changes. 

\begin{figure}[t]
  \centering
  \includegraphics[width=\linewidth]{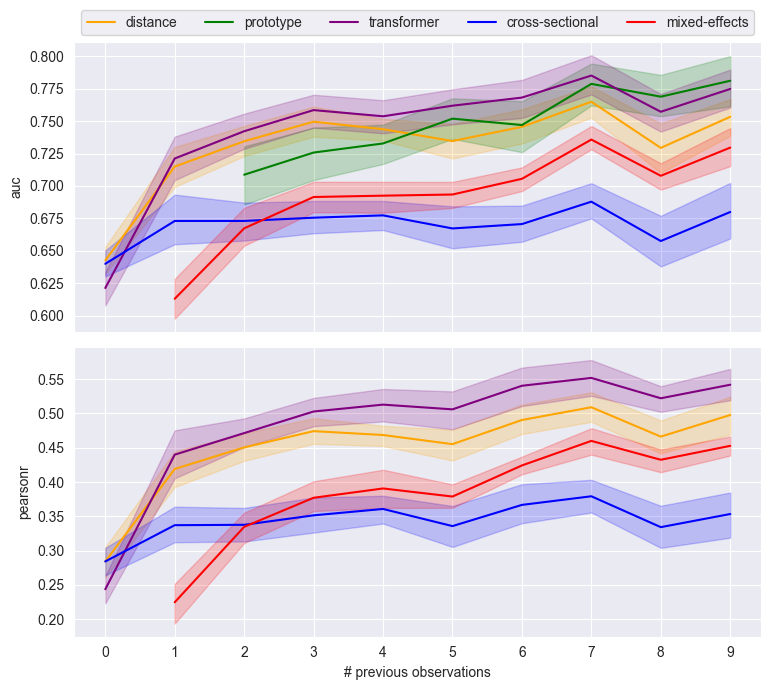}
  \caption{Performance comparison across participants for different numbers of previous observations (with 95\% CI).  Top panel: classification - AUC. Bottom panel: regression - Pearson correlation.}
  \label{fig:metrics_curve}
\end{figure}

\noindent
\textbf{Prototypical Network:} We implement a metric-learning approach following \cite{snell2017prototypical}. We learn a projection function $g: \mathbb{R}^{1024} \rightarrow \mathbb{R}^{64}$ (implemented as three linear layers with ReLU activations and hidden dimension of 32) that maps speech embeddings to a space where similar fatigue states cluster together. 

For the classification task, we compute class prototypes from the support set:
\begin{equation}
c^i_k = \frac{1}{|S^i_{t,k}|}\sum_{j \in S^i_{t,k}}g(x^i_j)
\end{equation}
where $S^i_{t,k}$ contains previous observations from speaker $i$ and class $k$. Prototypes are therefore created for each speaker and fatigue state. Classification probabilities are then derived from Euclidean distances to these prototypes. While effective for classification, this approach does not naturally extend to regression, which is why we exclude it from our regression analysis.

\noindent
\textbf{Transformer-based Model:} Our final approach treats speaker adaptation as a sequence modelling problem, leveraging transformers' ability to learn from in-context examples \cite{brown2020language, yadlowsky2023pretrainingdatamixturesenable}. For each speaker, we construct sequences:
\begin{equation}
[x^i_0, y^i_0, x^i_1, y^i_1, ..., x^i_t]
\end{equation}
Using a GPT2 architecture (15 layers, 8 heads), we frame fatigue prediction as next-token prediction, computing loss only at target positions (the $y$s). To form the sequences, targets are augmented to match feature dimensionality by sampling each dimension from $\mathcal{N}(y^i_j, 0.1^2)$. We use a mean squared error loss for each dimension of the augmented target, which is then averaged to produce the final loss.

\section{Results}

\subsection{Model Performance}

\begin{table}[t]
\caption{Performance averaged for points with 6 and 7 previous observations. Dist, Proto, Tr, CS, ME respectively stand for Distance, Prototype, Transformer, Cross-sectional and Mixed-effects methods. bacc, rec, prec, rho stand for balanced accuracy, recall, precision and pearson correlation. Standard deviation is in bracket.}
\label{tab:perf_t6}
\setlength{\tabcolsep}{3pt}  % reduce column separation
\footnotesize   % reduce font size
\centering
\begin{tabular}{@{}lccccc@{}}  % remove edge spacing
\toprule
& \multicolumn{5}{c}{\textbf{Method}} \\
\cmidrule(lr){2-6}
& \textbf{Dist} & \textbf{Proto} & \textbf{Tr} & \textbf{CS} & \textbf{ME} \\
\midrule
\multicolumn{5}{l}{\textbf{Classification}}  \\
auc & .755[.034] & .763[.049] & .777[.040] & .679[.036] & .721[.020]  \\
bacc & .684[.041] & .694[.044] & .706[.034] & .627[.032] & .657[.018]  \\
rec & .671[.060] & .666[.071] & .690[.078] & .606[.057] & .630[.029]  \\
prec & .640[.058] & .664[.064] & .671[.056] & .582[.052] & .618[.027]  \\
\multicolumn{5}{l}{\textbf{Regression}} \\
rho & .500[.056] & n/a & .546[.069] & .373[.069] & .442[.028]  \\
RMSE & 5.32[.27] & n/a & 5.04[.31] & 5.59[.28] & 5.38[.09]  \\
\bottomrule
\end{tabular}
\end{table}

We observed that, overall, prediction accuracy increases with the number of available observations per speaker (Figure \ref{fig:metrics_curve}). The cross-sectional baseline, having no speaker adaptation, maintains constant performance (AUC $\sim$ 0.68, correlation $\sim$ 0.37). The prototype model and mixed-effects approaches require at least one or two previous observations to make predictions, as they need examples to build speaker-specific representations.

The meta-learning approaches demonstrate notable improvements as more speaker observations become available, surpassing both the cross-sectional and mixed-effects baselines. The transformer model performs best, reaching an AUC of 0.78 and a correlation of 0.55 with a minimum of 6 prior observations (Table \ref{tab:perf_t6}). Compared to conventional approaches, this represents improvements of 14\% in AUC and 46\% in correlation versus cross-sectional modeling, and 8\% and 23\% respectively versus mixed-effects.
The transformer approach also offers practical advantages. Unlike the prototype model, it handles both classification and regression. Unlike the distance model, it doesn't require ensembling predictions from all previous observations. And critically, unlike both alternatives, it provides `cold start' predictions for speakers with no previous observations, making it more flexible in deployment.

\subsection{Sensitivity Analysis}
\label{subsection:sensitivity_analysis}

\noindent
\textbf{Sequence Effect Control:} Given twice-daily recordings (pre/post shift), participants' fatigue patterns could show periodicity that models might exploit without learning from speech content. To test possible impacts of this effect, we created a control experiment using a null model that maintains our transformer architecture but replaces speech features with dataset-wide averages. Testing on periodic (alternating high/low fatigue), balanced (fixed-length), and random sequences, we found the model achieves high performance (AUC=0.922) only when both trained and tested on periodic sequences (Table \ref{tab:null}). Performance drops to chance level with random ordering. Based on these findings, we use random sequence ordering in all previous experiments to ensure performance gains reflect learning from speech content rather than temporal patterns.

\begin{figure}[t]
  \centering
  \includegraphics[width=\linewidth]{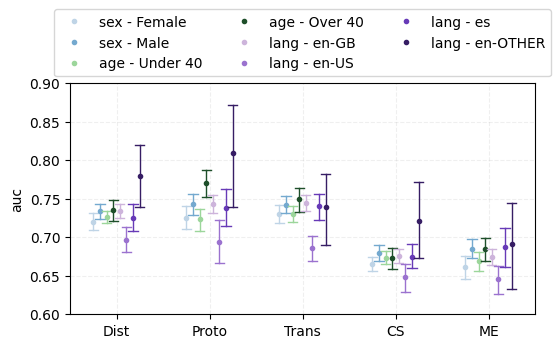}
  \caption{AUC for points with up to 7 previous observations for different demographics (with 95\% CI)}
  \label{fig:auc_bias}
\end{figure}

\noindent
\textbf{Fairness Analysis:} To assess potential biases, we first analysed the dataset using confound-aware metrics \cite{polle2024revealing}, finding balanced representation across demographic groups (confound-aware AUC $\sim$ 0.51 for sex, age, and language). However, examining model performance across these groups reveals some concerning patterns:
\begin{itemize}
\item Performance is consistently lower for female speakers across all methods, with this disparity most pronounced in the mixed-effects approach.
\item Speakers under 40 show lower performance in most methods except cross-sectional, with the prototype model showing particularly strong age-related bias.
\item While British and Spanish speakers show similar performance, US English speakers have markedly lower accuracy across all methods.
\end{itemize}
Figure \ref{fig:auc_bias} shows these disparities in detail, averaging performance across observations 0-7 to capture early-stage adaptation. While all methods show some demographic biases, the prototype network appears to amplify these differences more than other approaches as evidenced by the wider spread in AUC across demographic groups for `Proto'. This suggests that while the transformer achieves the best overall performance, there remains important work to be done in ensuring equitable performance across demographic groups.

\begin{table}[t]
  \caption{Null model average AUC on different sequence types. The standard deviation is in brackets.}
  \label{tab:null}
  \footnotesize 
  \centering
  \begin{tabular}{l c c c}
    \toprule
    & \multicolumn{3}{c}{\textbf{Tested on}} \\
    \cmidrule(lr){2-4}
    \textbf{Trained on} & \textbf{Periodic} & \textbf{Balanced} & \textbf{Random} \\
    \midrule
    Periodic & $.922$ $[.090]$ & $.440$ $[.046]$ & $.498$ $[048]$ \\
    Balanced & $.664$ $[.186]$ & $.751$ $[.123]$ & $.506$ $[.037]$  \\
    Random & $.519$ $[.047]$ & $.515$ $[.036]$ &  $.502$ $[.029]$ \\
    \bottomrule
  \end{tabular}
\end{table}

\section{Conclusions and Future Work}

In this work, we reformulate speaker-dependent fatigue monitoring as a meta-learning problem to eliminate the need for constant model retraining. We demonstrate that meta-learning approaches can effectively create personalized fatigue detection models for individual users while maintaining practical feasibility for real-world applications. Our transformer-based approach achieves strong performance (Pearson correlation of 0.55, AUC of 0.78) for individuals with at least 6 previous labelled speech recordings. This represents substantial improvements over traditional approaches (46\% over cross-sectional, 23\% over mixed-effects) while offering key practical advantages: 1) no need for continuous retraining, 2) ability to handle both regression and classification, and 3) availability of predictions even without previous speaker observations.

Our findings have important implications for real-world deployment. The requirement of only a few initial interactions with users to achieve personalisation makes it suitable for health monitoring applications. Users would need to report their fatigue levels for just a few days before receiving fully automated, personalised predictions from speech alone – a considerable advancement in making fatigue monitoring both accurate and practical.

Some limitations of our approach should be noted. While our models show promising overall performance -- particularly given the variability in devices and recording settings -- we observe performance variations across demographic groups, with differences in accuracy for female speakers and those under 40.  Our investigation is also limited to the time since sleep variable, which is only one component of fatigue. Additionally, our current analysis uses fixed-length speech samples from standardised reading tasks. The same dataset includes other speech activities such as image descriptions and question-answering tasks, offering opportunities to explore more naturalistic speech patterns. Future work could investigate how our approach generalises across these different speech tasks, potentially enabling more flexible real-world applications that don't rely on specific text reading.

Importantly, beyond fatigue (operationalised as time since sleep), this approach could be adapted to other mental states or health symptoms where personalisation is crucial. Particularly interesting would be applications to conditions with slower temporal dynamics, such as depression or chronic pain, where collecting diverse samples may be more challenging, but the need for personalised monitoring is equally important. Research could also focus on making these models more robust across demographic groups through targeted data collection or architectural modifications.

Our work demonstrates that meta-learning is feasible for personalised health monitoring from speech while highlighting important considerations for moving such systems from research to real-world deployment. The performances achieved with just a few calibration samples per user suggest this approach could impact how we monitor fatigue and potentially other health conditions in real-world settings.

\bibliographystyle{IEEEtran}
\bibliography{main}

\end{document}